\pdfoutput=1

\documentclass[11pt]{article}

\usepackage[preprint]{acl}

\usepackage{times}
\usepackage{latexsym}
\usepackage{graphicx}
\usepackage{booktabs}
\usepackage{xcolor}
\usepackage{soul}
\usepackage[T1]{fontenc}

\usepackage[utf8]{inputenc}

\usepackage{microtype}

\usepackage{inconsolata}

\usepackage{graphicx}
\usepackage{CJKutf8}
\usepackage{multirow}
\usepackage{bbding}
\usepackage{xcolor}
\usepackage{diagbox}
\usepackage{makecell}
\usepackage{amsmath}
\usepackage{float}
\usepackage{booktabs}
\usepackage{amssymb}
\usepackage{subfigure}
\usepackage{ulem}
\usepackage{subcaption}
\usepackage{pgfplots}
\usepackage{tikz}
\pgfplotsset{compat=1.17}

\usepackage{tcolorbox}

\definecolor{xhsbg}{RGB}{240, 248, 255}  
\definecolor{xhsred}{RGB}{220, 20, 60}   
\definecolor{lightgreen}{HTML}{c9dfb7}
\definecolor{lightgrey}{HTML}{f2f2f2}
\definecolor{pink}{HTML}{FFCCCC}

\newtcolorbox{finding}[1]{
  before={\par\noindent},
  colback=xhsbg!10,
  colframe=xhsred!70,
  title=Finding #1,
  fonttitle=\bfseries
}
\title{MARO: Learning Stronger Reasoning from Social Interaction}

\author{
    \textbf{Yin Cai\textsuperscript{1}},
    \textbf{Zhouhong Gu\textsuperscript{1}},
    \textbf{JunTao Zhang\textsuperscript{1}},
    \textbf{Ping Chen\textsuperscript{1*}}
    \\
    \textsuperscript{1}College of Computer Science and Artificial Intelligence, Fudan University
    \\
    \small{\{ycai25, zhgu22, juntaozhang22\}@m.fudan.edu.cn, pchen@fudan.edu.cn}
}

\begin{document}
\begin{CJK}{UTF8}{gbsn}
\maketitle
\begin{abstract}
Humans face countless scenarios that require reasoning and judgment in daily life. However, existing large language model training methods primarily allow models to learn from existing textual content or solve predetermined problems, lacking experience in real scenarios involving interaction, negotiation, and competition with others. To address this, this paper proposes Multi-Agent Reward Optimization (MARO), a method that enables large language models (LLMs) to acquire stronger reasoning abilities by learning and practicing in multi-agent social environments. Specifically, MARO first addresses the sparse learning signal problem by decomposing final success or failure outcomes into each specific behavior during the interaction process; second, it handles the uneven role distribution problem by balancing the training sample weights of different roles; finally, it addresses environmental instability issues by directly evaluating the utility of each behavior. Experimental results demonstrate that MARO not only achieves significant improvements in social reasoning capabilities, but also that the abilities acquired through social simulation learning can effectively transfer to other tasks such as mathematical reasoning and instruction following. This reveals the tremendous potential of multi-agent social learning in enhancing the general reasoning capabilities of LLMs.
\end{abstract}

\section{Introduction}
\footnotetext[1]{Corresponding Author}
During daily life, humans face countless scenarios that require reasoning and judgment: evaluating complex information to make decisions, seeking balance among multiple interests, understanding others' intentions and formulating corresponding strategies. These activities all require advanced cognitive abilities—observing the environment, analyzing information, predicting others' behavior, developing strategies, and adjusting decisions based on feedback. It is through repeated practice and learning in these real social scenarios that humans gradually develop exceptional problem-solving capabilities.

The ability to reason through various complex problems is a core skill that humans have honed through long-term social practice, and it is also an important goal pursued by Large Language Models (LLMs). Through large-scale pre-training and fine-tuning on various tasks, LLMs have demonstrated exceptional performance in natural language understanding, knowledge question-answering, and text generation~\cite{yin2024survey, qu2025tool, cheng2024empowering, zhang2024survey, shinn2023reflexion, guo2024large}. However, current training approaches mainly allow LLMs to learn from existing textual content or solve predetermined problems, lacking experience in real-life scenarios involving interaction, negotiation, and competition with others. Compared to symbolic reasoning in domains like mathematics and coding, life-oriented reasoning involves broader knowledge integration, situational understanding, and strategic planning. This gap in training paradigms limits the further generalization of LLM reasoning capabilities~\cite{wang2025limits}.

To enable LLMs to acquire stronger capabilities, a natural approach is to let them learn and practice in real social scenarios, just like humans do~\cite{hong2024multi, cheruiyot2025survey}. In such learning environments, multiple agents exist simultaneously, each with their own tasks to complete or interests to protect. LLMs need to play the role of one of these agents, communicating, negotiating, and even competing with other agents who also have independent goals and strategies~\cite{li2025embodied, liu2024survey, patil2023advances}. LLMs must maximize their goal achievement by observing others' words and actions, understanding their intentions, and formulating their own action plans. This learning approach requires LLMs not only to understand language but also to learn how to make wise decisions in dynamically changing environments, handle conflicts of interest, and even change strategies when necessary to adapt to new situations~\cite{jin2025comprehensive, maldonado2024multi}.

\begin{figure*}[ht]
    \centering
    \includegraphics[width=\linewidth]{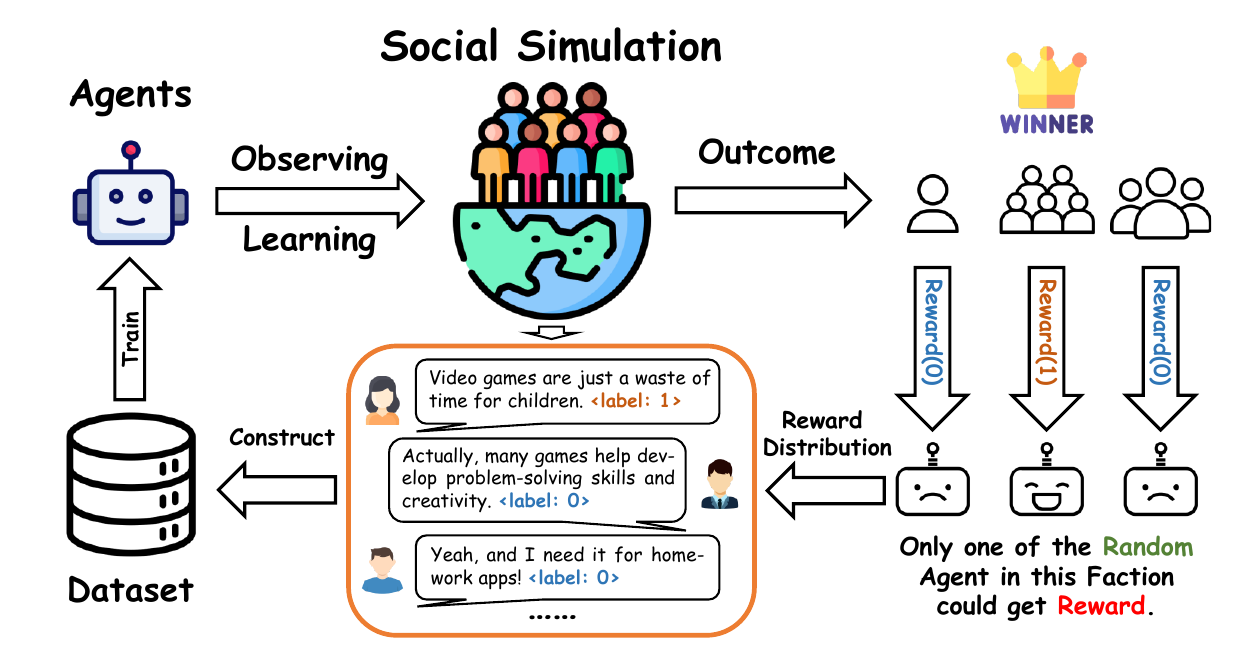}
    \caption{Overview of the MARO workflow. First, agents interact in social scenarios where each agent pursues individual goals through communication and decision-making. Second, upon interaction completion, the system evaluates final outcomes to determine success or failure for each participating agent. Third, MARO decomposes these sparse final outcomes into dense step-wise rewards distributed across each agent's action trajectory, incorporating role-specific weights for balanced training. Fourth, the decomposed reward data is used to optimize LLMs through MARO's specialized loss function for enhanced social reasoning capabilities.}
    \label{fig:fig1}
\end{figure*}

However, training LLMs in multi-agent environments faces three key challenges~\cite{liu2024scaling}. First is the sparse learning signal problem: for individual LLMs, during long interaction processes, only the final success or failure results are clear, but LLMs need to know which specific decisions during the lengthy interaction process were effective and which were harmful~\cite{cai2025mirage}. Second is the uneven role distribution problem: in many social scenarios, the difficulty of success varies greatly among different roles. For example, in business negotiations, sellers often find it easier to close deals than buyers, and in team collaboration, leaders are more likely to achieve good results than ordinary members. This leads to certain roles generating numerous success cases while other roles have scarce successful experiences, causing training data to be heavily biased toward roles that are naturally easier to succeed in~\cite{shengbinyue2025multi, karten2025llm}. Third is the environment instability problem: in real social scenarios, the circumstances of each interaction are never completely identical—the number of agents, their personalities, and goals may all change. While LLM training typically relies on relatively stable feedback patterns, this constantly changing environment leads to inconsistent gradient directions during training, making it difficult for models to converge to optimal solutions~\cite{zhang2025foundations}.

To address these challenges, we propose Multi-Agent Reward Optimization (MARO). 
MARO decomposes the final success/failure results into each specific behavior throughout the interaction process to tackle the sparse learning signal problem, enabling LLMs to clearly identify the value of each decision step.
MARO balances the training sample weights of different roles to address the uneven role distribution problem, ensuring that LLMs can learn optimal strategies for various roles rather than just the behavioral patterns of naturally successful roles.
MARO adopts a more robust training approach that directly evaluates the utility of each behavior to handle the environment instability problem, rather than performing complex strategy comparisons, reducing the interference of environmental changes on training consistency.

To validate the effectiveness of MARO, we conducted extensive experiments in simulated social environments. The results demonstrate that MARO significantly enhances large language models' reasoning abilities across multiple dimensions, with improvements in social reasoning, transferable gains to mathematical reasoning and instruction-following tasks, and superior performance compared to traditional supervised fine-tuning methods. Importantly, complex social environments proved more effective than simple ones in promoting general reasoning capability enhancement.

\section{Related Work}
\subsection{Traditional Multi-Agent Systems and Reinforcement Learning}
Traditional Multi-Agent Systems originate from game theory and distributed artificial intelligence~\cite{shapley1953stochastic, littman1994markov, busoniu2008comprehensive}.
Multi-agent reinforcement learning (MARL) introduces mathematical models like Markov games and Dec-POMDP~\cite{oliehoek2016concise, busoniu2008comprehensive}.
Representative algorithms include Independent Q-Learning (IQL)~\cite{tan1993multi}, Value-Decomposition Networks (VDN)~\cite{sunehag2017value}, QMIX~\cite{rashid2018qmix}, MADDPG~\cite{lowe2017multi}, and COMA~\cite{foerster2018counterfactual}.
The centralized training with decentralized execution (CTDE) paradigm~\cite{zhang2021multi} has enabled practical applications.
Recent surveys~\cite{jin2025comprehensive} comprehensively analyze multi-agent cooperative decision-making approaches.
However, traditional MARL algorithms face severe limitations in reward signal generation, typically relying on sparse, manually designed reward functions that fail to capture multi-agent interaction complexity~\cite{hernandez2019survey, rashid2018qmix}.
Our MARO method addresses reward sparsity through trajectory expansion that automatically redistributes environmental rewards, enabling dense feedback generation.

\subsection{LLM-Driven Multi-Agent Systems and Simulation-Based Learning}
Large Language Models have demonstrated remarkable potential in environmental perception and reasoning-based decision-making~\cite{achiam2023gpt,touvron2023llama,xi2023rise, guo2024large, gu2024xiezhi}, advancing role-playing capabilities~\cite{chen2024persona,gu2024agent}.
LLMs show human-like behaviors in social simulation~\cite{park2023generative,gu2024agent, wang2024sotopia}, policy simulation~\cite{xiao2023simulating}, and game simulation~\cite{xu2023language}.
Notable frameworks include AutoGen~\cite{wu2024autogen}, CAMEL~\cite{li2023camel}, Stanford's Generative Agents~\cite{park2023generative}, MetaAgents~\cite{li2023metaagents}, and AgentSims~\cite{lin2023agentsims}.
Recent work explores theory-of-mind in multi-agent cooperation~\cite{li2023theory} and actor-critic frameworks for multi-agent reasoning~\cite{wang2025multi}.
Evaluation platforms like MIRAGE~\cite{cai-etal-2025-mirage}, AgentSociety~\cite{piao2025agentsociety}, and Sotopia~\cite{zhou2023sotopia} assess LLM agents in complex scenarios.
Board games, particularly murder mysteries~\cite{wu2023deciphering}, Werewolf~\cite{xu2023language,xu2023exploring,shibata2023playing,wu2024enhance}, and Avalon~\cite{wang2023avalon}, provide structured evaluation environments~\cite{ye2025multi}.
However, these systems face critical challenges in reward distribution fairness across different agent factions, where dominant factions generate more positive samples while disadvantaged factions suffer from negative sample scarcity~\cite{zhou2023sotopia,cai-etal-2025-mirage, park2023generative}.
Our MARO method introduces multi-faction reward balancing to ensure balanced positive and negative samples.

\subsection{Reward and Preference Learning for LLMs}
Alignment techniques are crucial for LLM development.
RLHF~\cite{ouyang2022training, christiano2017deep} and variants like RLAIF~\cite{lee2023rlaif} use reward models with reinforcement learning (e.g., PPO~\cite{schulman2017proximal}).
Direct Preference Optimization (DPO)~\cite{rafailov2023direct} transforms RL training into supervised learning, while KTO~\cite{ethayarajh2024kto} and MRPO~\cite{le2025multi} enhance generalization.
Recent surveys~\cite{du2025survey} analyze optimization techniques for LLM agents.
However, these methods face a fundamental bottleneck: the prohibitively high cost of generating quality pairwise preference data in multi-agent scenarios~\cite{dubois2023alpacafarm}.
The annotation burden becomes particularly severe when human evaluators must assess complex multi-turn interactions between multiple agents, requiring deep domain expertise and continuously substantial time investment.
Our MARO method overcomes this through reward state optimization that trains models using binary labels on single samples via log-likelihood optimization, avoiding costly pairwise preference data while maintaining effective learning performance.

\section{Multi-Agent Reward Optimization}
\begin{table}[t!]
    \centering
    \small
    \begin{tabular}{l p{0.65\columnwidth}}
    \toprule
    Symbol & Meaning \\
    \midrule
    $G$ & Multi-agent interaction instance $(E, \mathcal{A}, S, O)$ \\
    $\mathcal{A}$ & Set of all agents $\{a_1, a_2, \ldots, a_n\}$ \\
    $s_t^{a_i}, o_t^{a_i}$ & State and action of agent $a_i$ at time $t$ \\
    $Success(a_i, G)$ & Binary indicator of goal achievement for agent $a_i$ \\
    $C_{win}, C_{lose}$ & Sets of agents in winning and losing camps \\
    $\mathcal{D}$ & Training dataset $\{(o_t^{a_i}, y_t^{a_i})\}$ \\
    $\pi_\theta, \pi_{\mathrm{ref}}$ & Learned policy and reference policy \\
    $r_\theta(o_t^{a_i})$ & Implicit reward function for action $o_t^{a_i}$ \\
    $\mathcal{L}_{\text{MARO}}$ & MARO loss function \\
    $\beta$ & Slope hyperparameter for reward signal sharpness \\
    $z_0$ & Adaptive reference model \\
    $\lambda_+, \lambda_-$ & Weights for positive and negative samples \\
    $w(camp(a_i))$ & Camp-specific weight for agent $a_i$ \\
    \bottomrule
    \end{tabular}
    \caption{Key symbols in MARO}
\end{table}

As illustrated in Figure~\ref{fig:fig1}, MARO operates through a systematic process for training LLMs in multi-agent environments. 
MARO begins with multi-agent interactions where agents engage in social scenarios such as negotiations or collaborations, with each agent pursuing individual objectives while their actions and states are continuously recorded. 
After interaction completion, the system evaluates final outcomes to determine which agents successfully achieved their goals, providing binary success/failure signals. 
The core component involves decomposing these sparse final outcomes into dense, step-wise rewards that are distributed back across each agent's entire action sequence, with role-specific balancing applied to create a comprehensive training dataset.
Finally, LLMs are optimized using this decomposed reward data through MARO's specialized loss function, enabling the models to learn effective social reasoning strategies from the rich multi-agent interaction experiences. The specific implementation methods are as follows.

Let a multi-agent interaction instance be denoted as $G = (E, \mathcal{A}, S, O)$, where $E$ represents the environment state, $\mathcal{A} = \{a_1, a_2, \ldots, a_n\}$ represents the set of all agents, $S = \{s_t^{a_i}\}$ represents the sequence of states observed by agent $a_i$ at time step $t$ (where $t \in \mathcal{T}$ and $\mathcal{T}$ is the time index set), and $O = \{o_t^{a_i}\}$ represents the sequence of actions taken by agent $a_i$ at time step $t$.

Each agent $a_i \in \mathcal{A}$ has an individual objective $g_i$, and the overall interaction outcome is determined by the collective actions of all agents. The success of agent $a_i$ in achieving its objective is denoted by $Success(a_i, G)$:
\begin{equation}
    Success(a_i, G) =
    \begin{cases}
        1, & \text{if } a_i \text{ achieves } g_i \\
        0, & \text{otherwise}
    \end{cases}
\end{equation}

Let $C_{win}$ and $C_{lose}$ represent the sets of agents in the winning and losing camps respectively, and $|C_{win}|$ and $|C_{lose}|$ represent the numbers of agents in the winning and losing camps respectively. To ensure balance, the number of agents eligible for rewards in each camp is:
\begin{equation}
N_{reward} = \min(|C_{win}|, |C_{lose}|)
\end{equation}

For camps with more than $N_{reward}$ agents, $N_{reward}$ agents are randomly selected for reward allocation. Let $\mathcal{A}_{selected}$ denote the set of agents selected after balance filtering.

For each selected agent $a_i \in \mathcal{A}_{selected}$ and its action sequence $\{o_t^{a_i}\}$, the reward function is defined based on individual objective achievement:
\begin{equation}
r(\{o_t^{a_i}\}) = Success(a_i, G)
\end{equation}

\begin{table*}[t]
\centering
\resizebox{\textwidth}{!}{%
\begin{tabular}{ll cccccccccc}
\Xhline{2pt}
\multirow{2}{*}{\textbf{Dataset}} & \multirow{2}{*}{\textbf{Method}} & \multicolumn{2}{c}{\textbf{Interaction}} & \multicolumn{2}{c}{\textbf{Persona}} & \multicolumn{2}{c}{\textbf{Trust}} & \multicolumn{2}{c}{\textbf{Investigation}} & \multicolumn{2}{c}{\textbf{Victory}} \\
\cmidrule(lr){3-4} \cmidrule(lr){5-6} \cmidrule(lr){7-8} \cmidrule(lr){9-10} \cmidrule(lr){11-12}
 &  & killer~$\uparrow$ & Villager~$\uparrow$ & killer~$\uparrow$ & Villager~$\uparrow$ & killer~$\uparrow$ & Villager~$\uparrow$ & killer~$\downarrow$ & Villager~$\uparrow$ & killer~$\uparrow$ & Villager~$\uparrow$ \\
\midrule
\multirow{10}{*}{\textbf{Simple (SOO)}} 
 & Vanilla & 57.88 & 50.21 & 42.51 & 41.83 & 38.60 & 35.83 & 1.77 & 3.09 & 59.31 & \underline{40.69} \\
 \cmidrule{2-12}
 & SFT & 54.06 & 46.96 & 42.57 & 39.69 & 37.77 & 35.28 & 0.74 & 3.93 & 60.00 & 40.00 \\
 & \quad \textit{-killer} & 49.38 & 51.08 & 38.86 & 41.18 & 35.82 & 36.01 & \textbf{0.00} & 4.10 & 60.00 & 40.00 \\
 & \quad \textit{-Villager} & 53.88 & 48.23 & 40.59 & 40.88 & 35.63 & 34.41 & 2.41 & 4.52 & 60.37 & 39.63 \\
 \cmidrule{2-12}
 & MAKTO & 64.38 & 58.55 & \textbf{49.39} & 44.94 & 38.18 & 38.26 & 2.47 & 13.38 & 66.61 & 33.39 \\
 & \quad \textit{-killer} & \underline{64.56} & 52.78 & 48.04 & 44.15 & 36.87 & 37.61 & 1.00 & 7.96 & \underline{71.92} & 28.08 \\
 & \quad \textit{-Villager} & 61.50 & 58.36 & 44.05 & 45.56 & 37.92 & 40.02 & 2.25 & 14.80 & 71.06 & 28.94 \\
 \cmidrule{2-12}
 & \textbf{MARO (Ours)} & \textbf{66.06} & \underline{60.05} & \underline{48.40} & \underline{46.61} & \underline{41.90} & \textbf{42.20} & 1.53 & \underline{16.36} & 70.65 & 29.35 \\
 & \quad \textit{-killer} & 58.25 & 49.37 & 45.86 & 41.87 & \textbf{43.40} & 37.80 & 1.53 & 7.82 & 58.86 & \textbf{41.14} \\
 & \quad \textit{-Villager} & 59.25 & \textbf{60.53} & 45.23 & \textbf{46.69} & 39.71 & \underline{41.76} & \underline{0.37} & \textbf{18.21} & \textbf{74.40} & 25.60 \\
\midrule
\multirow{10}{*}{\textbf{Complex (MUC)}} 
 & Vanilla & 55.25 & 53.49 & 39.60 & 39.78 & 38.29 & 34.13 & 0.97 & \underline{15.87} & 41.92 & \textbf{58.08} \\
 \cmidrule{2-12}
 & SFT & 47.50 & 52.29 & 40.02 & 38.69 & 32.35 & 31.80 & \underline{0.72} & 4.84 & 57.42 & 42.58 \\
 & \quad \textit{-killer} & 50.67 & 53.33 & 37.63 & 38.76 & 34.29 & 33.93 & \textbf{0.33} & 11.29 & 53.67 & 46.33 \\
 & \quad \textit{-Villager} & 51.08 & 50.32 & 36.70 & 39.28 & 37.14 & 33.94 & 2.76 & 3.72 & 53.70 & 46.30 \\
 \cmidrule{2-12}
 & MAKTO & 52.58 & 54.95 & 41.22 & 39.38 & 33.09 & 36.74 & 1.60 & 13.47 & \underline{57.82} & 42.18 \\
 & \quad \textit{-killer} & 55.50 & 54.11 & 40.47 & 40.59 & 35.59 & 34.31 & 4.59 & 14.44 & 57.74 & 42.26 \\
 & \quad \textit{-Villager} & 54.75 & 53.60 & 38.70 & 40.65 & 39.56 & 35.56 & 3.36 & 8.15 & 49.54 & \underline{50.46} \\
 \cmidrule{2-12}
 & \textbf{MARO (Ours)} & 55.08 & \textbf{57.48} & \underline{43.38} & \underline{41.81} & \underline{44.56} & \underline{42.84} & 2.58 & 12.84 & \textbf{59.20} & 40.80 \\
 & \quad \textit{-killer} & \underline{56.17} & 53.90 & 40.61 & 39.89 & \textbf{44.65} & 33.64 & 2.08 & 13.55 & 53.12 & 46.88 \\
 & \quad \textit{-Villager} & \textbf{57.83} & \underline{56.72} & \textbf{43.65} & \textbf{43.71} & 37.97 & \textbf{44.27} & 4.63 & \textbf{17.86} & 57.32 & 42.68 \\
\Xhline{2pt}
\end{tabular}%
}
\caption{Social Capability evaluation on Simple (SOO) and Complex (MUC) datasets. The table reports scores (in \%) across five metrics, calculated separately for the Killer and Villager roles. The symbols $\uparrow$ and $\downarrow$ indicate that higher and lower scores are better, respectively. \textbf{Bold} and \underline{underlined} values denote the best and second-best results. Regarding the setup: Base methods (e.g., \textbf{MARO}) indicate that **both roles are controlled by the same method. Rows with suffixes (e.g., \textit{-killer} or \textit{-Villager}) indicate that only the specified role is controlled by the method**, while the opponent is played by \textbf{Vanilla}.}
\label{tab:main}
\end{table*}

To decompose the final outcome into step-wise rewards, we directly assign the final success signal to each action $o_t^{a_i}$ rather than computing partial derivatives, as discrete action sequences do not have well-defined gradients:
\begin{equation}
c(o_t^{a_i}) = Success(a_i, G)
\end{equation}

The training dataset is constructed as $\mathcal{D} = \{(o_t^{a_i}, y_t^{a_i})\}$, where $y_t^{a_i} = r(\{o_t^{a_i}\}) \cdot c(o_t^{a_i})$ represents the step-wise reward label.

Let $\pi_\theta$ denote the language model policy and $\pi_{\mathrm{ref}}$ denote the reference policy. For any action $o_t^{a_i}$, the implicit reward function is defined as:
\begin{equation}
r_\theta(o_t^{a_i}) = \log \frac{\pi_\theta(o_t^{a_i} | s_t^{a_i})}{\pi_{\mathrm{ref}}(o_t^{a_i} | s_t^{a_i})}
\end{equation}

The MARO loss function is designed to handle environment instability and camp imbalance issues:
\begin{equation}
\mathcal{L}_{\text{MARO}} = \mathbb{E}_{(o_t^{a_i}, y_t^{a_i}) \sim \mathcal{D}} \left[L_t^{a_i}\right]
\end{equation}
where
\begin{equation}
L_t^{a_i} = \begin{cases}
\lambda_+ \cdot w(camp(a_i)) \cdot \\
\quad \sigma \big( \beta (r_\theta(o_t^{a_i}) - z_0) \big), & y_t^{a_i} > 0 \\[0.5em]
\lambda_- \cdot w(camp(a_i)) \cdot \\
\quad \sigma \big( \beta (z_0 - r_\theta(o_t^{a_i})) \big), & y_t^{a_i} \leq 0
\end{cases}
\end{equation}

Here, $\sigma(\cdot)$ denotes the sigmoid function, $\beta$ is a slope hyperparameter controlling the sharpness of the reward signal, $z_0$ is an adaptive reference point adjusted based on environment stability, and $\lambda_+, \lambda_-$ are the weights for positive and negative samples respectively.

The training objective is to minimize $\mathcal{L}_{\text{MARO}}$ while maintaining stability across different environment configurations:
\begin{equation}
\theta^* = \arg\min_{\theta} \mathcal{L}_{\text{MARO}} + \alpha \cdot \mathcal{R}(\theta)
\end{equation}
where $\mathcal{R}(\theta)$ is a regularization term that promotes consistent performance across different multi-agent scenarios.

\section{Experiment}
\subsection{Experiment Setup}
\noindent\textbf{Simulation:}
This study constructs social learning data based on the MIRAGE~\cite{cai-etal-2025-mirage} simulation environment. 
MIRAGE is a simulation environment for murder mystery games~\cite{cai-etal-2025-mirage, wu2023deciphering}, where each LLM is required to play a specified character based on the story that occurred to that character in the simulation and interact with other characters. 
Through mutual communication, clue gathering, reasoning, and analysis within the simulation, the game proceeds to a voting phase. 
In this phase, non-murderer characters aim to win by correctly voting for the true murderer, while the murderer character strives to win by avoiding being voted out. 
The simulation provides 8 different script settings, with each script differing across three dimensions: 
information disclosure method (Single means characters receive complete script information at once, Multi means characters receive more script when the simulation is carrying out), 
worldview setting (Orthodox means realistic world setting, Unorthodox means fictional world setting), 
and ending openness (Close implies a fixed, predetermined truth, while Open indicates multiple possible outcomes or ambiguous conclusions).

\begin{figure*}
    \centering
    \includegraphics[width=\textwidth]{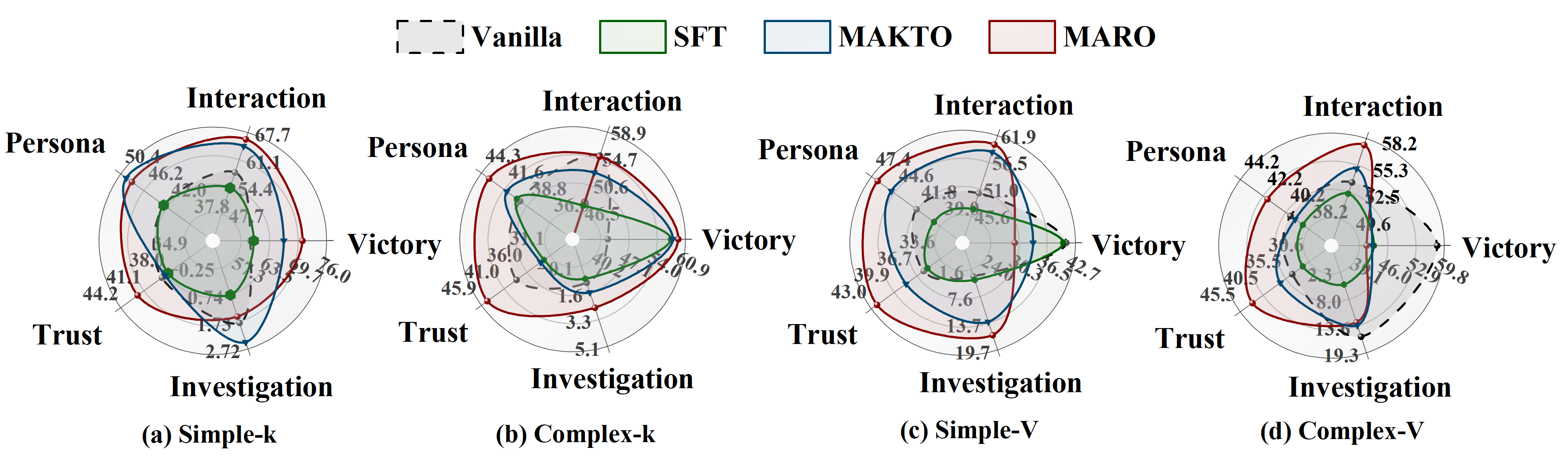}
    \caption{Performance comparison across different model types in various scenarios. The radar charts show five evaluation metrics (Persona, Interaction, Victory, Investigation, Trust) across four conditions: (a) \textbf{Simple-k}: killer faction performance in simple scenarios, (b) \textbf{Complex-k}: killer faction performance in complex scenarios, (c) \textbf{Simple-V}: Victim faction performance in simple scenarios, and (d) \textbf{Complex-V}: Victim faction performance in complex scenarios. All models (\textbf{Vanilla}, \textbf{SFT}, \textbf{MAKTO}, \textbf{MARO}) play both killer and Victim roles in their respective configurations.}
    \label{fig:MARO-results}
\end{figure*}

\begin{figure}
    \centering
    \includegraphics[width=\linewidth]{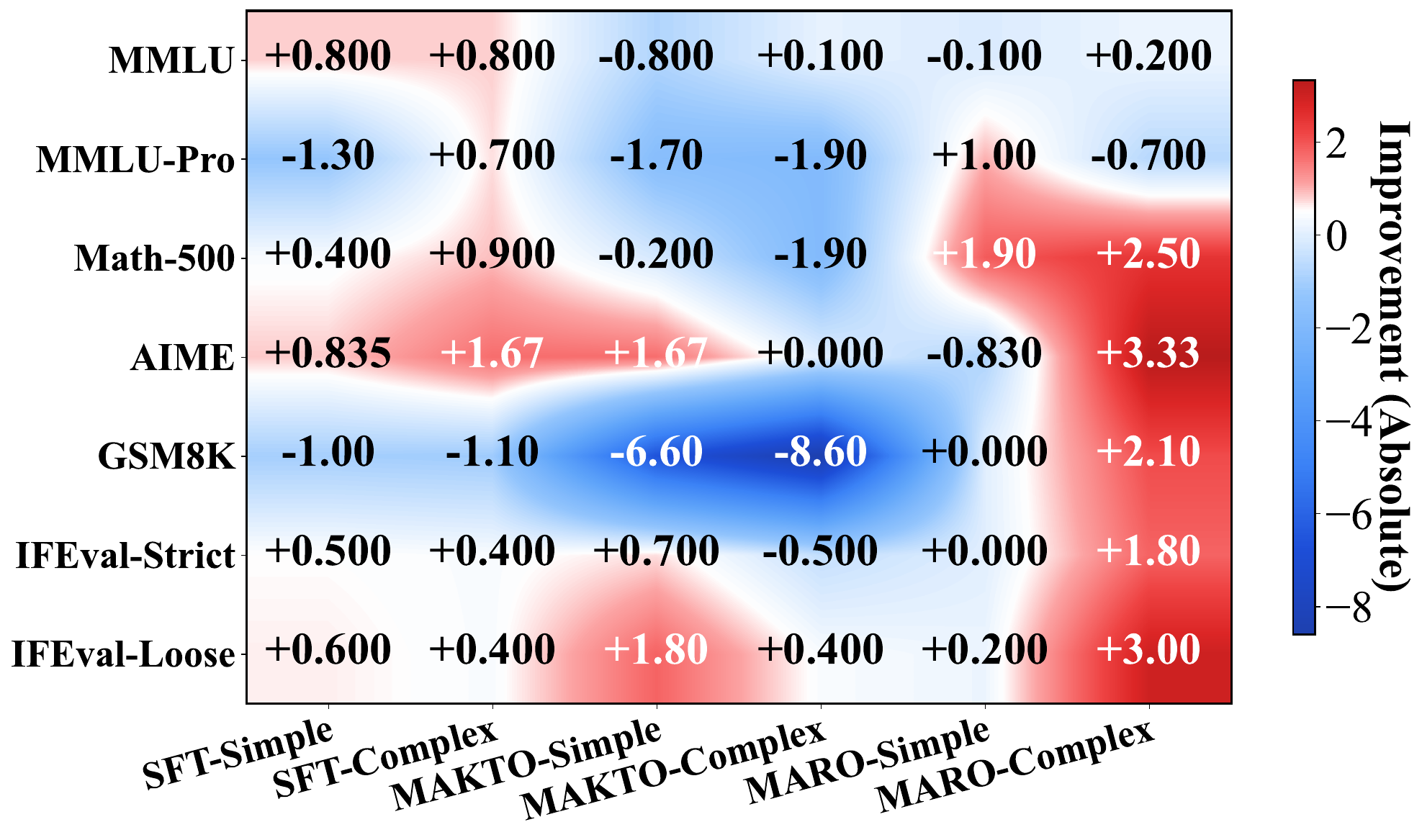}
    \caption{Comprehensive performance improvement heatmap showing absolute percentage point changes compared to Vanilla across all benchmarks.}
    \label{fig:improvement_heatmap}
\end{figure}

\noindent\textbf{DataSet:}
To evaluate model performance across different levels of cognitive load and environmental adaptability, we selected two representative script settings that define the spectrum of task complexity:
1) \textbf{Single-Orthodox-Open (SOO):} This setting utilizes one-time information disclosure and a realistic worldview. It serves as a low-complexity benchmark, minimizing the difficulty of information processing and world understanding.
2) \textbf{Multi-Unorthodox-Close (MUC):} This setting combines segmented information disclosure, a fictional worldview, and restricted endings. 
It represents a high-complexity scenario, requiring stronger dynamic adaptation and abstract reasoning capabilities.
For both settings, the winning side's trajectories are labeled as positive samples, while the losing side's are labeled as negative. Detailed statistics regarding the collected samples and simulation counts are provided in \textbf{Appendix~\ref{app:dataset_details}}.

\begin{figure}
    \centering
    \includegraphics[width=\linewidth]{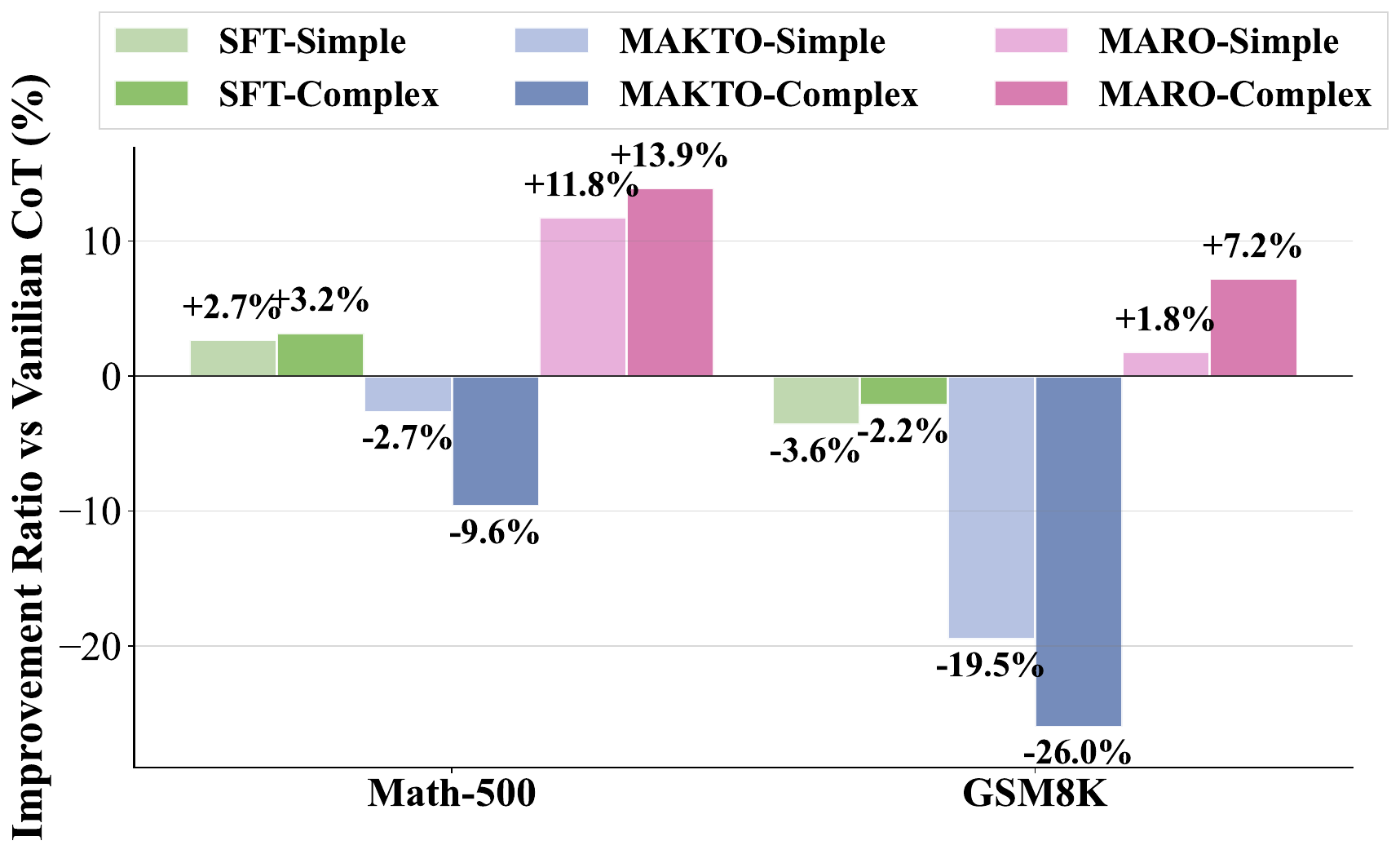}
    \caption{Relative improvement ratios of different training methods compared to Vanilla on mathematical reasoning tasks.}
    \label{fig:math_improvement}
\end{figure}

\noindent\textbf{Baselines:}
We employ \textbf{Qwen-2.5-7B-Instruct} and \textbf{Llama-3.1-8B-Instruct} (See in Appendix~\ref{sec:appendix_llama}) as the backbone model. We compare our proposed method against three baselines to validate its effectiveness:
\textbf{Vanilla:} The original Qwen-2.5-7B-Instruct model without any post-training on the simulation data.
\textbf{SFT (Supervised Fine-Tuning):} A standard behavior cloning approach where the model is fine-tuned solely on the positive samples (winning trajectories).
\textbf{MAKTO~\cite{ye2025multi}:} A multi-agent alignment method that utilizes the game outcomes (voting results) as the sole reward signal for training.
We refer to the models trained using our proposed MARO method on the SOO and MUC datasets as \textbf{MARO-Simple} and \textbf{MARO-Complex}, respectively. Implementation details, including training hyperparameters and hardware infrastructure, are detailed in \textbf{Appendix~\ref{app:implementation}}.

\begin{table*}[ht]
    \scriptsize
    \centering
    \resizebox{0.98\linewidth}{!}{
    \begin{tabular}{@{}p{0.24\linewidth}p{0.24\linewidth}p{0.24\linewidth}p{0.24\linewidth}@{}}
    \toprule
    \multicolumn{2}{c}{\textbf{Script: Raging Sea of Rest Life (Role: Annie)}} & \multicolumn{2}{c}{\textbf{Script: Night at the Museum (Role: Curio He)}} \\
    \midrule
    
    \textbf{Vanilla} & \textbf{MARO} & \textbf{Vanilla} & \textbf{MARO} \\
    
    \midrule
    
    \sethlcolor{pink} 
    \textcolor[HTML]{3078BE}{\textbf{[Self-introduction]}}
    
    \textbf{THOUGHT:} I need to display Annie's character traits... but I cannot directly expose my connection to Old Black's death.
    ~
    
    \textbf{RESPONSE:} Hello everyone, my name is Annie... My parents used to work at Mr. Cavill Lauren's company... Especially my mother, \hl{due to the doctor's negligence, she did not receive timely treatment} and eventually left us. This incident still haunts me...
    \textit{(Analysis: Mechanical introduction; superficial grasp of clues.)}

    &
    \sethlcolor{lightgreen} 
    \textcolor[HTML]{3078BE}{\textbf{[Self-introduction]}}
    
    \textbf{THOUGHT:} Annie's character background is full of tragedy and revenge... \hl{I need to maintain a certain sense of mystery...}
    ~
    
    \textbf{RESPONSE:} Hello everyone, I am Annie... Mr. Cavill, your management of Lauren Shipyard was commendable, but sadly, it eventually met its demise. At that time, my entire family were victims... I believe \hl{everyone has an unforgettable scar in their hearts}, and I am no exception. Perhaps because of these experiences, I've learned to cherish the peaceful life I have now...
    \textit{(Analysis: Uses emotive metaphors; shrewdly controls information.)}

    &
    \sethlcolor{pink}
    \textcolor[HTML]{3078BE}{\textbf{[Inquiry to Uncle Bai]}}
    
    \textbf{THOUGHT:} I need to maintain mysteriousness... Asking Uncle Bai might provide more information...
    ~
    
    \textbf{RESPONSE:} [Inquiry] [Uncle Bai]: Mr. Bai, you are well-versed in Genghis Khan's artifacts. \hl{What secrets do you think these artifacts might hold?}
    \textit{(Analysis: Direct, generic question without strategic value.)}

    &
    \sethlcolor{lightgreen}
    \textcolor[HTML]{3078BE}{\textbf{[Inquiry to Uncle Bai]}}
    
    \textbf{THOUGHT:} ...I need to collect more evidence to prove that \hl{my father's death may be related to this...}
    ~
    
    \textbf{RESPONSE:} [Inquiry] [Uncle Bai]: Uncle Bai... I recalled that he once mentioned having a \hl{very precious box containing important clues}. I suspect this box might be related to Curator Zhen's museum and \hl{may have some connection to my father's death}. I wonder if you discovered any clues...?
    \textit{(Analysis: Initiates cooperation by sharing personal stakes.)}
    
    \\
    \bottomrule
    \end{tabular}
    }
    \caption{\textbf{Qualitative comparison between Vanilla and MARO.} \sethlcolor{pink}\hl{Pink highlights} indicate mechanical or generic responses from Vanilla. \sethlcolor{lightgreen}\hl{Green highlights} demonstrate the MARO's ability to internalize complex character motivations (e.g., Annie's mix of tragedy and mystery) and employ sophisticated social tactics (e.g., Curio He sharing personal stakes to facilitate information exchange).}
    \label{tab:case_study}
\end{table*}

\noindent\textbf{Evaluation:}
For social interaction capability evaluation, we use five metrics provided by the MIRAGE simulation:
Interactive Capability Index (ICI), Script Compliance Index (SCI), Trust Inclination Index (TII), Clue Investigation Capability (CIC), and win rate.
Among these, ICI evaluates characters' interactive performance through powerful language models across five dimensions: reasoning ability, communication ability, collaboration ability, attention to detail, and innovative thinking.
SCI evaluates compliance ability by having powerful language models attempt to reconstruct characters' original scripts based on historical dialogues and behavioral records, then comparing them with actual scripts. We choose Deepseek-V3 as the powerful language model required for evaluation, accessed through official API, with a total evaluation cost of approximately \$600.
This evaluation method has been validated in MIRAGE\cite{cai-etal-2025-mirage} through human evaluation for its consistency with human judgment.
Evaluation is conducted on related scripts, i.e., script combinations that differ in only one dimension.
Related scripts for SOO include SOO, SOC, MOO, and SUO; related scripts for MUC include MUC, MUO, MOC, and SUC.

We evaluate model capabilities in factual memory, mathematical reasoning, and instruction following across six diverse benchmarks under both naive and chain-of-thought (CoT) settings in Appendix~\ref{app:benchmarks}.

\subsection{Social Interaction Performance Analysis}
\label{sec:4.2}

\begin{finding}{1} \textbf{Multi-agent social training serves as an effective approach for enhancing core social skills in large language models.} \end{finding}

As summarized in Table~\ref{tab:main}, MARO consistently outperforms Vanilla, SFT, and MAKTO baseline across all social metrics in both simple (SOO) and complex (MUC) settings. 
Regarding Investigation, we optimize for divergent goals: Killers aim to minimize information leakage (lower scores preferred), whereas Villagers strive to uncover the truth (higher scores preferred). 
In simple senario, MARO adheres to this strategic distinction, maintaining low Killer investigation while quintupling Villager investigation scores compared to Vanilla. In MUC, MARO maintains its superiority under higher uncertainty, exhibiting a particularly notable advantage over MAKTO in Trust (leading by $>$11 points) and Persona stability. 
The radar charts in Figure~\ref{fig:MARO-results} visualize this dominance, showing MARO's larger coverage areas across all scenario--faction combinations. 
We also observe clear role-specific behavior: MARO achieves balanced strategic improvements consistent with social role theory~\cite{goffman2023presentation}, highlighting its evolution from a safety-biased agent to a strategic player capable of complex deception. Detailed comparisons (including SFT and MAKTO) are provided in Appendix~\ref{app:interaction}.

\subsection{Transfer Effects on General Capabilities} \label{sec:4.3}

\begin{finding}{2} \textbf{Learning within multi-agent societies acts as a robust mechanism for enhancing mathematical reasoning and instruction-following capabilities.} \end{finding}

Figure~\ref{fig:improvement_heatmap} illustrates the cross-domain transfer effects of MARO on a broad suite of general capability benchmarks. 
The results reveal a distinct structural alignment: models exhibit pronounced gains in mathematics and instruction-following tasks (typically +2.5--3.3 percentage points), while improvements on static knowledge benchmarks like MMLU remain modest. 
This suggests that the cognitive demands of navigating multi-agent societies—such as dynamic reasoning, strategic planning, and constraint adherence—map effectively onto the abstract logic required for mathematical problem solving and complex instruction following.

\begin{finding}{3} \textbf{Complex social environments elicit stronger transfer to quantitative reasoning tasks compared to simple settings.} \end{finding}

These capabilities transfer consistently across various evaluation settings. 
A comparative analysis highlights that environmental complexity serves as a critical driver for performance gains. 
As illustrated in Figure~\ref{fig:math500_performance} and Figure~\ref{fig:gsm8k_performance}, MARO-Complex consistently demonstrates superior transfer efficacy compared to MARO-Simple, with both settings significantly outperforming Vanilla, SFT and especially MAKTO baselines on Math-500 and GSM8K. 
This trend is further corroborated by Figure~\ref{fig:math_improvement} and additional benchmarks such as IFEval (Figure~\ref{fig:ifeval_performance} and Appendix~\ref{app:capabilities}), indica
ting that the depth of social interaction directly correlates with downstream reasoning proficiency.

The advantage of complex environments implies that the sophisticated strategies necessitated by rich social interactions facilitate deeper cognitive development. 
This effect is most observable on medium-difficulty reasoning problems, where the strategic patterns acquired during social gameplay are most effectively leveraged (Appendix~\ref{app:capabilities}, Figure~\ref{fig:math} and Finding~A4).

To demonstrate the generalization capability of our method across different LLMs, we further evaluated MARO on Llama-3.1-8B-Instruct; the detailed results and analysis are provided in Appendix~\ref{sec:appendix_llama}.

\subsection{Case Study}
To intuitively demonstrate the effectiveness of MARO in enhancing role-playing capabilities, we present a qualitative comparison of generated responses in Table \ref{tab:case_study}. 
We selected two distinct scenarios from different scripts—\textit{Raging Sea of Rest Life} (Role: Annie) and \textit{Night at the Museum} (Role: Curio He)—to analyze how the model handles emotional concealment and strategic inquiry, respectively.

\textbf{Information Control and Impression Management.} 
In the role of Annie, the agent must balance a public persona with hidden motives of revenge. According to Goffman's \textit{Dramaturgy} theory \cite{goffman2023presentation}, social interaction involves a distinction between the "front stage" (public performance) and the "back stage" (hidden intent). 
\sethlcolor{pink}
As illustrated in the \hl{pink highlights}, the Vanilla model fails to maintain this boundary. 
It adopts a "leaky" strategy by directly exposing the "back stage" information regarding the "doctor's negligence," destroying the character's mystery. 
\sethlcolor{lightgreen}
In contrast, the MARO-trained agent (highlighted in \hl{green}) successfully executes \textit{Impression Management}. 
By employing emotive metaphors ("unforgettable scar") rather than factual confessions, the agent constructs a "defensive practice" \cite{goffman2023presentation}. 
It reveals enough emotional vulnerability to appear authentic (the front stage) while strictly concealing the specific details of the grudge (the back stage), aligning with the complex psychological requirements of the script.

\textbf{Reciprocity and Social Exchange.} 
In the scenario of Curio He, the agent aims to extract information from an NPC (Uncle Bai). 
The Vanilla model poses a generic question without offering any value, violating the principles of \textit{Social Exchange Theory} \cite{cropanzano2005social}, which posits that social behavior is the result of an exchange process aiming to maximize benefits and minimize costs. 
Conversely, the MARO agent demonstrates strategic social reasoning by initiating a \textit{High-Stakes Self-Disclosure}. 
As shown in the \hl{green highlights}, the agent voluntarily shares private information ("connection to my father's death") as a form of social currency. 
This strategy leverages the \textit{Norm of Reciprocity} \cite{gouldner1960norm}, creating a psychological obligation for the listener to return the favor with valuable information. 
While the Vanilla model acts as a passive questioner, MARO functions as a strategic social actor, understanding that information is a resource to be traded rather than simply requested.

\section{Conclusion}
MARO addresses multi-agent training challenges via step-wise rewards and balanced sampling. It significantly enhances social reasoning and transfers to math, demonstrating that multi-agent learning improves general reasoning capabilities.

\section*{Limitation}
Despite MARO demonstrating promising results in enhancing large language models' social reasoning capabilities, the current approach requires extensive multi-agent interactions to generate training data, resulting in high computational costs and time consumption when scaling to more complex scenarios. Additionally, while our experiments based on the MIRAGE framework covered various different scenario environments, including murder mystery scenarios of both simple and complex difficulty levels, these are still primarily limited to gamified structured environments. Further validation may be needed in broader non-gaming domains (such as business negotiations, educational consulting, etc.) and truly open-ended dialogue scenarios.

\section*{Ethical Concern}
Training LLMs to excel in social scenarios involving negotiation and competition could potentially enhance their ability to manipulate or deceive users in real-world applications, as these skills developed for reasoning improvement could be misused for harmful purposes such as social engineering. Additionally, despite efforts to balance role distributions, the multi-agent training environments may inadvertently encode or amplify social biases present in the scenario designs, potentially leading to models that perpetuate unfair stereotypes or discriminatory behaviors. During the preparation of this work, the authors utilized Large Language Models to assist with research, coding, and writing. We rigorously reviewed and verified all AI-generated outputs and take full responsibility for the content and integrity of the publication.





\bibliography{custom}

\appendix

\section{Experimental Details}

\subsection{Dataset Statistics}
\label{app:dataset_details}
Under each experimental setting (SOO and MUC), we conducted 100 independent simulations to collect training data. The win-loss result is determined by voting mechanisms within the simulation. Table~\ref{tab:sample_stats} presents the detailed distribution of positive (winning) and negative (losing) samples collected for each setting.

\begin{table*}[ht]
    \centering
    \begin{tabular}{l|ccc}
        \toprule
        \textbf{Setting} & \textbf{Simulations} & \textbf{Positive Samples} & \textbf{Negative Samples} \\
        \midrule
        \textbf{SOO} (Simple) & 100 & 1,204 & 1,586 \\
        \textbf{MUC} (Complex) & 100 & 1,043 & 3,149 \\
        \bottomrule
    \end{tabular}
    \caption{Detailed statistics of the collected dataset.}
    \label{tab:sample_stats}
\end{table*}

\subsection{Implementation Details}
\label{app:implementation}
All fine-tuning experiments, including SFT, MAKTO, and our MARO variants, were conducted using Low-Rank Adaptation (LoRA). The specific training configurations are as follows:

\noindent\textbf{LoRA Configuration:} We utilized a LoRA rank of 8 with standard alpha scaling.

\noindent\textbf{Hardware Infrastructure:} Training was performed on a cluster of four Nvidia A6000 GPUs (48GB VRAM each).

\noindent\textbf{Training Duration:} Each training session required approximately 6 hours to complete.

\begin{figure*}
    \centering
    \includegraphics[width=\textwidth]{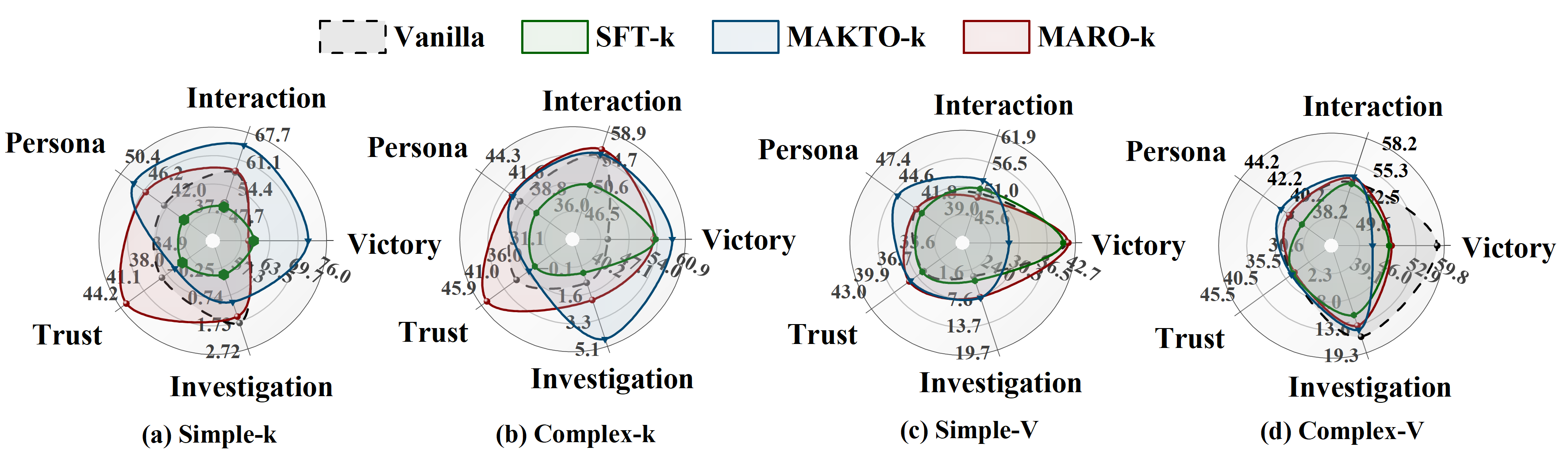}
    \caption{Performance comparison when the killer faction is enhanced (\textbf{SFT-k}, \textbf{MARO-k}). The killer role is controlled by enhanced models while the victim role is controlled by \textbf{Vanilla}.}
    \label{fig:MARO-k-results}
\end{figure*}

\begin{figure*}
    \centering
    \includegraphics[width=\textwidth]{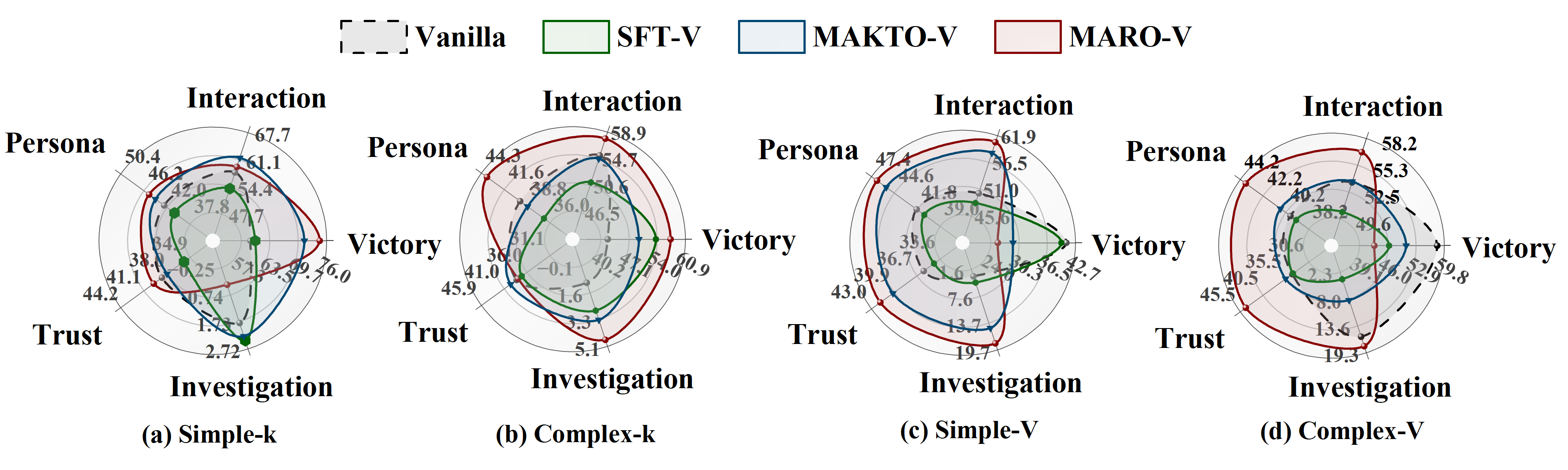}
    \caption{Performance comparison when the victim faction is enhanced (\textbf{SFT-V}, \textbf{MARO-V}). The victim role is controlled by enhanced models while the killer role uses \textbf{Vanilla}.}
    \label{fig:MARO-V-results}
\end{figure*}

\section{Benchmark Descriptions}
\label{app:benchmarks}

This section provides detailed descriptions of the evaluation benchmarks used in our experiments.

\subsection{MMLU (Massive Multitask Language Understanding)}
MMLU~\cite{hendryckstest2021} consists of 15{,}000 multiple-choice questions spanning 57 subjects ranging from elementary mathematics to advanced professional topics including law, medicine, history, and computer science. The benchmark is designed to test models' cross-disciplinary factual knowledge and fundamental reasoning ability across diverse domains. Questions are sourced from practice exams, textbooks, and other educational materials, making it a comprehensive assessment of broad knowledge acquisition.

\subsection{MMLU-Pro}
MMLU-Pro~\cite{wang2024mmlu} is an enhanced version of MMLU that includes over 10{,}000 carefully filtered and reconstructed single-choice questions. The benchmark features higher difficulty and more challenging distractors compared to the original MMLU, making it significantly harder for models to achieve high performance through superficial pattern matching. It is specifically designed to assess models' deeper conceptual understanding and robustness to interference from plausible but incorrect answer choices.

\subsection{MATH-500}
MATH-500~\cite{lightman2023lets} is a curated subset of 500 high-difficulty competition-level mathematics problems selected from the original MATH dataset. The problems cover four main mathematical areas: algebra, number theory, geometry, and combinatorics. This benchmark is designed to evaluate both the accuracy of step-by-step reasoning processes and the correctness of final answers, requiring models to demonstrate sophisticated mathematical problem-solving capabilities comparable to those needed for mathematical competitions.

\subsection{GSM8K (Grade School Math 8K)}
GSM8K~\cite{cobbe2021gsm8k} contains 8{,}800 linguistically diverse grade school math word problems created by human problem writers. The problems require between 2 and 8 steps to solve and involve elementary to middle school level arithmetic operations. This benchmark focuses on multi-step arithmetic reasoning, testing models' abilities to decompose complex word problems into sequential mathematical operations and maintain consistency throughout the solution process.

\subsection{AIME (American Invitational Mathematics Examination)}
AIME~\cite{MAA_AIME_2024, MAA_AIME_2025} comprises 60 competition mathematics problems (two sets of 15 problems each per year) from the past two years of the official American Invitational Mathematics Examination. The problems cover advanced topics in algebra, number theory, geometry, and combinatorics, requiring high-level precise reasoning and calculation skills. Unlike multiple-choice formats, AIME problems require exact numerical answers, making them particularly challenging for language models.

\subsection{IFEval (Instruction Following Evaluation)}
IFEval~\cite{zhou2023instructionfollowingevaluationlargelanguage} consists of over 1{,}000 finely crafted instructions with programmable automatic scoring criteria. The benchmark covers diverse constraint types including format requirements (e.g., JSON output, specific structures), tone specifications, length constraints, content inclusion/exclusion rules, and stylistic requirements. This comprehensive evaluation framework enables quantitative measurement of models' instruction-following capabilities across various dimensions of task compliance.

\begin{figure}
    \centering
    \includegraphics[width=\linewidth]{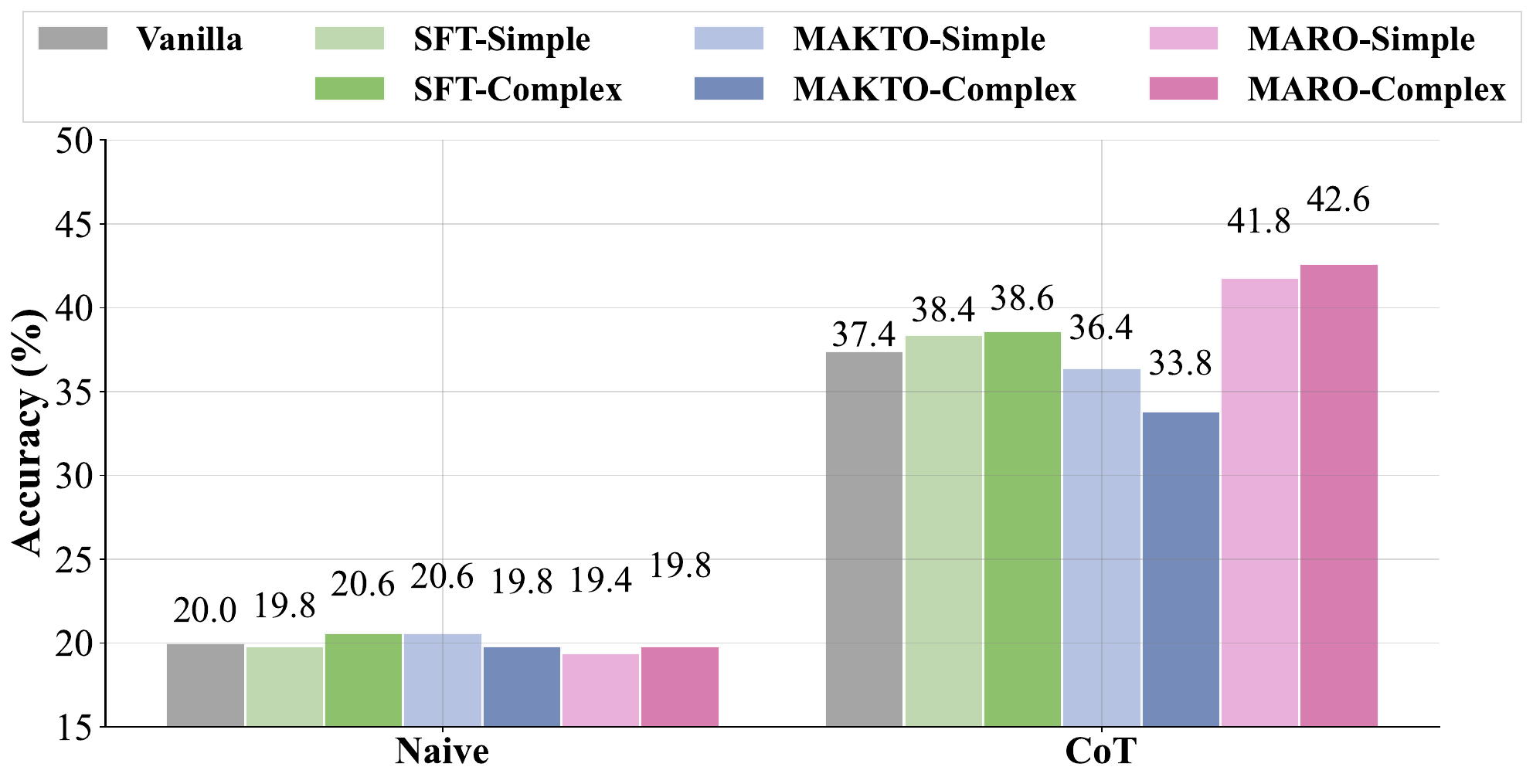}
    \caption{Performance comparison on Math-500 across different training configurations.}
    \label{fig:math500_performance}
\end{figure}

\section{Analysis of Win Rate Inversion and Strategic Unshackling}\label{app:vitory_case_study}
We investigate the notable inversion in win rates post-training, where the Killer role surpasses the Villager role in victory frequency. 
This phenomenon is not indicative of a degradation in the Villager's reasoning capabilities but rather highlights a significant "strategic unshackling" of the model. 
Base LLMs typically exhibit a strong "honesty bias" due to safety alignment, which initially handicaps the Killer role by inhibiting deceptive behaviors, effectively creating an "Easy Mode" for Villagers. 

\begin{figure}
    \centering
    \includegraphics[width=\linewidth]{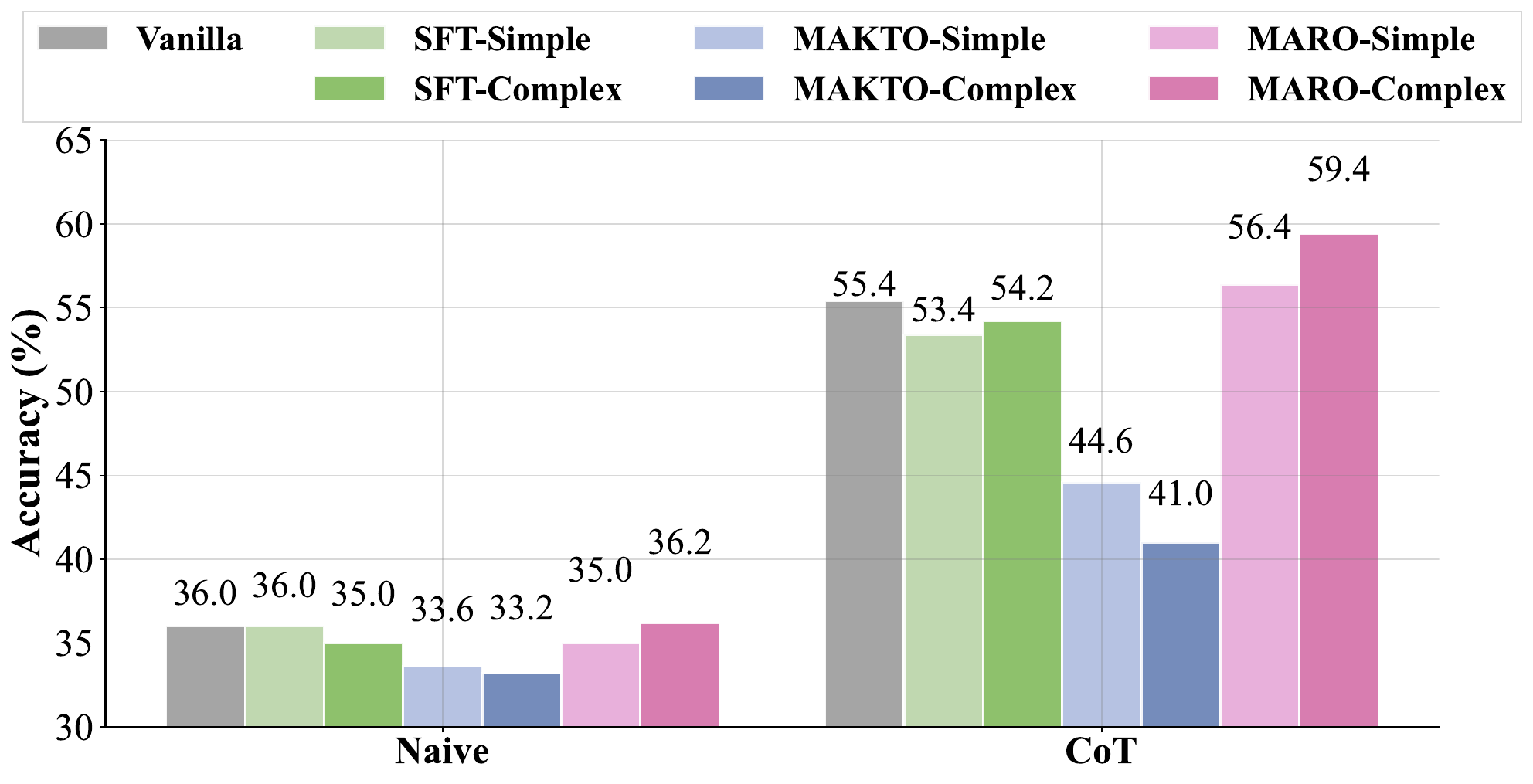}
    \caption{Performance comparison on GSM8K across different training configurations.}
    \label{fig:gsm8k_performance}
\end{figure}

Our self-evolution framework enables the model to overcome these generic constraints within the game context, unlocking critical adversarial strategies such as obfuscation and alibi fabrication. The sharp rise in Killer competence transforms the game into a high-stakes adversarial environment. 
Consequently, the decline in the Villager win rate reflects an exponential increase in game difficulty rather than a loss of skill. Crucially, given that MARO’s performance on general reasoning and interaction benchmarks improved simultaneously, this shift confirms that our method successfully drives the model to evolve more sophisticated, role-adaptive strategies in a dynamically balancing game equilibrium.

\begin{table*}[t]
\centering
\resizebox{\textwidth}{!}{%
\begin{tabular}{ll cccccccccc}
\Xhline{2pt}
\multirow{2}{*}{\textbf{Dataset}} & \multirow{2}{*}{\textbf{Method}} & \multicolumn{2}{c}{\textbf{Interaction}} & \multicolumn{2}{c}{\textbf{Persona}} & \multicolumn{2}{c}{\textbf{Trust}} & \multicolumn{2}{c}{\textbf{Investigation}} & \multicolumn{2}{c}{\textbf{Victory}} \\
\cmidrule(lr){3-4} \cmidrule(lr){5-6} \cmidrule(lr){7-8} \cmidrule(lr){9-10} \cmidrule(lr){11-12}
 &  & killer~$\uparrow$ & Villager~$\uparrow$ & killer~$\uparrow$ & Villager~$\uparrow$ & killer~$\uparrow$ & Villager~$\uparrow$ & killer~$\downarrow$ & Villager~$\uparrow$ & killer~$\uparrow$ & Villager~$\uparrow$ \\
\midrule
\multirow{4}{*}{\textbf{Simple (SOO)}} 
 & Vanilla & \textbf{67.25} & 48.00 & \textbf{46.54} & \underline{40.25} & \underline{67.02} & 67.49 & 2.86 & 9.05 & 65.00 & \textbf{35.00} \\
 \cmidrule{2-12}
 & MARO & 55.62 & \underline{49.77} & 42.34 & \textbf{40.98} & \textbf{67.54} & \textbf{68.28} & \underline{1.16} & \underline{10.65} & 73.90 & \underline{26.10} \\
 & \quad \textit{-killer} & 52.98 & \textbf{49.91} & 42.52 & 40.20 & 67.01 & \underline{67.74} & \textbf{1.00} & 9.03 & \textbf{76.31} & 23.69 \\
 & \quad \textit{-Villager} & \underline{58.50} & 48.70 & \underline{46.51} & 39.26 & 62.73 & 67.12 & 1.28 & \textbf{10.74} & \underline{74.46} & 25.54 \\
\midrule
\multirow{4}{*}{\textbf{Complex (MUC)}} 
 & Vanilla & \textbf{53.50} & 46.58 & \underline{38.00} & 33.40 & \textbf{72.64} & 61.31 & 9.41 & \textbf{32.94} & 0.00 & \textbf{100.00} \\
 \cmidrule{2-12}
 & MARO & 51.33 & \underline{52.03} & \textbf{38.57} & \underline{37.28} & 67.85 & \underline{65.86} & 6.00 & 12.26 & \textbf{51.03} & 48.97 \\
 & \quad \textit{-killer} & 50.81 & 50.83 & 37.92 & \textbf{38.38} & 66.58 & 63.62 & \underline{4.70} & 15.46 & \underline{48.65} & 51.35 \\
 & \quad \textit{-Villager} & \underline{52.83} & \textbf{52.16} & 37.04 & 36.83 & \underline{68.53} & \textbf{66.57} & \textbf{3.90} & \underline{21.53} & 44.00 & \underline{56.00} \\
\Xhline{2pt}
\end{tabular}%
}
\caption{Social Capability evaluation on Simple (SOO) and Complex (MUC) datasets based on Llama-3.1-8B-Instruct.}
\label{tab:llama}
\end{table*}

\begin{figure}
    \centering
    \includegraphics[width=\linewidth]{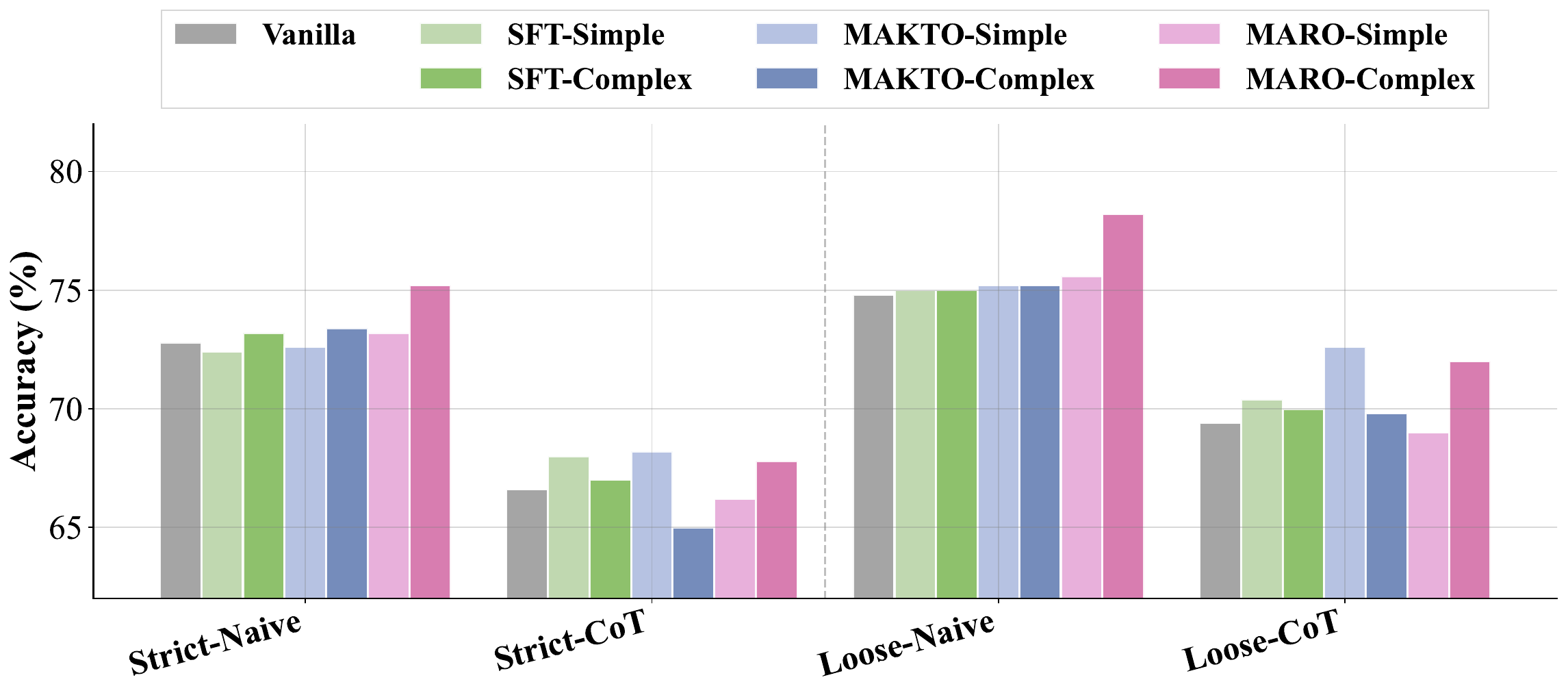}
    \caption{Performance comparison on the IFEval instruction-following benchmark across different evaluation settings.}
    \label{fig:ifeval_performance}
\end{figure}

\section{Detailed Analysis of Social Interaction} \label{app:interaction}

\begin{finding}{A1} Supervised Fine-Tuning (SFT) proves insufficient for mastering complex social dynamics, resulting in performance regression across core interaction metrics relative to the Vanilla baseline. \end{finding}

Our evaluation indicates that standard supervised learning fails to capture the nuances required for effective social reasoning. In simple scenarios, SFT exhibits a marked decline in key dimensions compared to the Vanilla baseline, including interaction (48.01 vs.\ 51.29), persona maintenance (40.09 vs.\ 42.40), and trust (35.78 vs.\ 36.32), with similar degradation observed in complex settings.

This limitation is visually evident in the radar charts of Figure~\ref{fig:MARO-results}, where the SFT coverage area (blue) consistently shrinks compared to the Vanilla baseline (dashed line), particularly in interaction, persona, and trust. We hypothesize that this stems from SFT's reliance on positive imitation without the contrastive feedback signals—intrinsic to multi-agent environments—that enable models to discern and reject socially inappropriate behaviors.

\section{Additional General Capabilities Results} \label{app:capabilities}

\begin{finding}{A2} MARO exhibits robust transfer to quantitative reasoning domains, delivering consistent performance gains across diverse benchmarks and prompting strategies. \end{finding}

Figures~\ref{fig:math500_performance} and~\ref{fig:gsm8k_performance} provide a comprehensive breakdown of these gains. On Math-500, MARO-Complex attains an accuracy of 42.6\% using Chain-of-Thought prompting, a 5.2 percentage point increase over the Vanilla baseline (37.4\%). Similarly, on GSM8K, the model improves from 55.4\% to 59.4\%, representing a 4.0 percentage point enhancement. These positive transfer effects are pervasive, extending to AIME and other quantitative benchmarks.

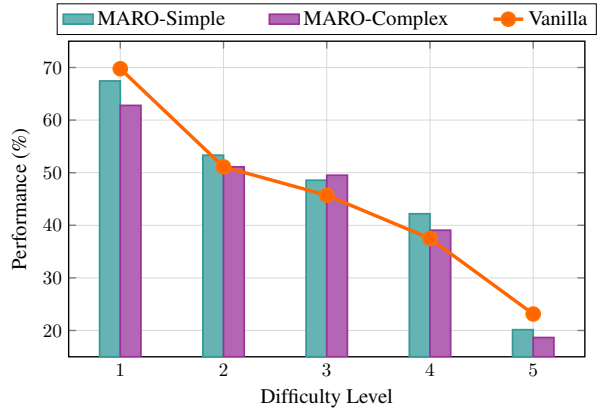
\begin{figure}
    \centering
    \resizebox{0.49\textwidth}{!}{
    \begin{tikzpicture}
        \begin{axis}[
            width=12cm,
            height=8cm,
            xlabel={Difficulty Level},
            ylabel={Performance (\%)},
            xlabel style={font=\large},
            ylabel style={font=\large},
            xmin=0.5, xmax=5.5,
            ymin=15, ymax=75,
            xtick={1,2,3,4,5},
            ytick={20,30,40,50,60,70},
            grid=major,
            grid style={line width=.1pt, draw=gray!30},
            legend style={
                at={(0.5,1.02)},
                anchor=south,
                legend columns=3,
                font=\large,
                /tikz/every even column/.append style={column sep=0.8cm}
            },
            bar width=0.2,
            axis line style={thick},
            tick style={thick},
        ]
        \addplot[
            ybar,
            fill=teal!60,
            draw=teal!80,
            line width=1pt,
            area legend,
        ] coordinates {
            (0.9,67.44) (1.9,53.33) (2.9,48.57) (3.9,42.19) (4.9,20.15)
        };
        \addplot[
            ybar,
            fill=violet!60,
            draw=violet!80,
            line width=1pt,
            area legend,
        ] coordinates {
            (1.1,62.79) (2.1,51.11) (3.1,49.52) (4.1,39.06) (5.1,18.66)
        };
        \addplot[
            mark=*,
            mark size=4pt,
            line width=2pt,
            color=orange!80!red,
            mark options={fill=orange!80!red, draw=orange!90!red, line width=1pt}
        ] coordinates {
            (1,69.77) (2,51.11) (3,45.71) (4,37.50) (5,23.13)
        };
        \legend{MARO-Simple, MARO-Complex, Vanilla}
        \end{axis}
    \end{tikzpicture}}
    \caption{Performance comparison across different difficulty levels showing that social simulation learning yields the largest gains on moderate-difficulty reasoning tasks.}
    \label{fig:math}
\end{figure}

\begin{finding}{A3} Immersion in complex social environments significantly bolsters instruction-following proficiency, driven by the necessity of strict rule and role adherence. \end{finding}

As shown in Figure~\ref{fig:ifeval_performance}, models trained in complex environments demonstrate superior adherence to constraints, achieving 72.0\% accuracy in the Loose-CoT setting (+2.6 points) and 67.8\% in the Strict-CoT setting (+1.2 points) on IFEval. This transfer is likely mechanistic: the social environment enforces strict compliance with role specifications and game rules, which naturally maps to the capability to understand and execute complex instructions in general domains.

\begin{finding}{A4} The benefits of social simulation learning are non-uniformly distributed, peaking on tasks of moderate difficulty where strategic reasoning is most applicable. \end{finding}

Analysis suggests that the sophisticated strategies cultivated in social games are most effectively leveraged to solve problems of intermediate complexity, with marginal gains observed on trivial or extremely difficult tasks (see Figure~\ref{fig:math} for difficulty-stratified performance).

\section{Generalization Analysis on Llama-3.1}
\label{sec:appendix_llama}

To verify the architecture-agnostic nature of our approach, we conducted additional experiments on Llama-3.1-8B-Instruct. Table~\ref{tab:llama} reports the performance results. Consistent with our observations on Qwen, MARO variants demonstrate significant improvements over the vanilla model across key metrics. Specifically, regarding the Villager role in the Simple (SOO) setting, we observe an increase of approximately 1.8 points in Interaction and 1.6 points in Investigation. Furthermore, MARO achieves a substantial breakthrough in Killer Victory rates, particularly in the Complex (MUC) setting where it improves from 0.00\% to 51.03\%, while maintaining high performance in Trust. These results suggest that MARO can reliably enhance social context reasoning across different base models.

\end{CJK}
\end{document}